\documentclass[10pt,twocolumn,letterpaper]{article}

\usepackage{iccv}              

\usepackage{multirow}

\definecolor{iccvblue}{rgb}{0.21,0.49,0.74}
\usepackage[pagebackref,breaklinks,colorlinks,allcolors=iccvblue]{hyperref}
\def\paperID{10064} 
\def\confName{ICCV}
\def\confYear{2025}

\title{Reasoning Physical Video Generation \\with Diffusion Timestep Tokens via Reinforcement Learning}

\author{Wang Lin$^{1,*}$, Liyu Jia$^{2,*}$, Wentao Hu$^{2,*}$, Kaihang Pan$^{1}$, Zhongqi Yue$^{2}$, \\Wei Zhao$^{3}$, Jingyuan Chen$^{1,\dagger}$, Fei Wu$^{1}$, Hanwang Zhang$^{2}$\\
$^1$Zhejiang University, $^2$Nanyang Technological University, $^{3}$Huawei Singapore Research Center
\\
{\tt\small linwanglw@zju.edu.cn},\quad
{\tt\small jingyuanchen@zju.edu.cn}
}

\usepackage[misc]{ifsym}
\newcommand\blfootnote[1]{
    \begingroup
    \renewcommand\thefootnote{}\footnote{#1}
    \addtocounter{footnote}{-1}
    \endgroup
}
\usepackage[hang]{footmisc}
\begin{document}
\maketitle

{
    \blfootnote{
        $^*$Equal contribution.\\
        $^\dagger$Corresponding author.\\
        }
}

\begin{abstract}
Despite recent progress in video generation, producing videos that adhere to physical laws remains a significant challenge.
Traditional diffusion-based methods struggle to extrapolate to unseen physical conditions (\textit{e.g.}, velocity) due to their reliance on data-driven approximations.    
To address this, we propose to integrate symbolic reasoning and reinforcement learning to enforce physical consistency in video generation. We first introduce the Diffusion Timestep Tokenizer (DDT), which learns discrete, recursive visual tokens by recovering visual attributes lost during the diffusion process.
The recursive visual tokens enable symbolic reasoning by a large language model.
Based on it, we propose the Phys-AR framework, which consists of two stages: The first stage uses supervised fine-tuning to transfer symbolic knowledge, while the second stage applies reinforcement learning to optimize the model’s reasoning abilities through reward functions based on physical conditions.    
Our approach allows the model to dynamically adjust and improve the physical properties of generated videos, ensuring adherence to physical laws.
Experimental results demonstrate that PhysAR can generate videos that are physically consistent.
\end{abstract}    
\section{Introduction}
\label{sec:intro}

Video generation~\cite{yang2024cogvideox,genmo2024mochi,opensora,kong2024hunyuanvideo,zhang2024mimicmotion} plays a pivotal role in driving advancements across various fields, from autonomous systems to innovative content creation.
However, a persistent challenge in this field lies in ensuring that generated videos adhere to fundamental physical law—such as motion continuity and collision dynamics—while maintaining visual realism.

As revealed by recent studies~\cite{liu2025generative,kang2024far}, diffusion-based approaches rely on iterative denoising processes that statistically approximate physical patterns through massive data fitting. While effective for interpolating within training distributions, this data-driven approximation fails to extrapolate to unseen physical conditions as it lacks explicit constraints derived from first-principle equations.

In practice, physical motion is simulated by analytical solving, \textit{i.e.}, by solving complex differential equations and deriving parametric principles of physics by stepwise deduction.
Recently, the advent of large language models (LLMs) like DeepSeek-R1~\cite{guo2025deepseek} and OpenAI’s o1~\cite{jaech2024openai} has demonstrated unprecedented symbolic reasoning capabilities~\cite{gao2024embedding,guan2025rstar}.
This suggests a promising paradigm: if video generation can be reformulated as a symbol-based reasoning process, then it is possible to reason about strict physical constraints through a large language model and thereby generate videos with consistent physical laws.

\begin{figure}[t]
\centering
\includegraphics[width=0.43\textwidth]{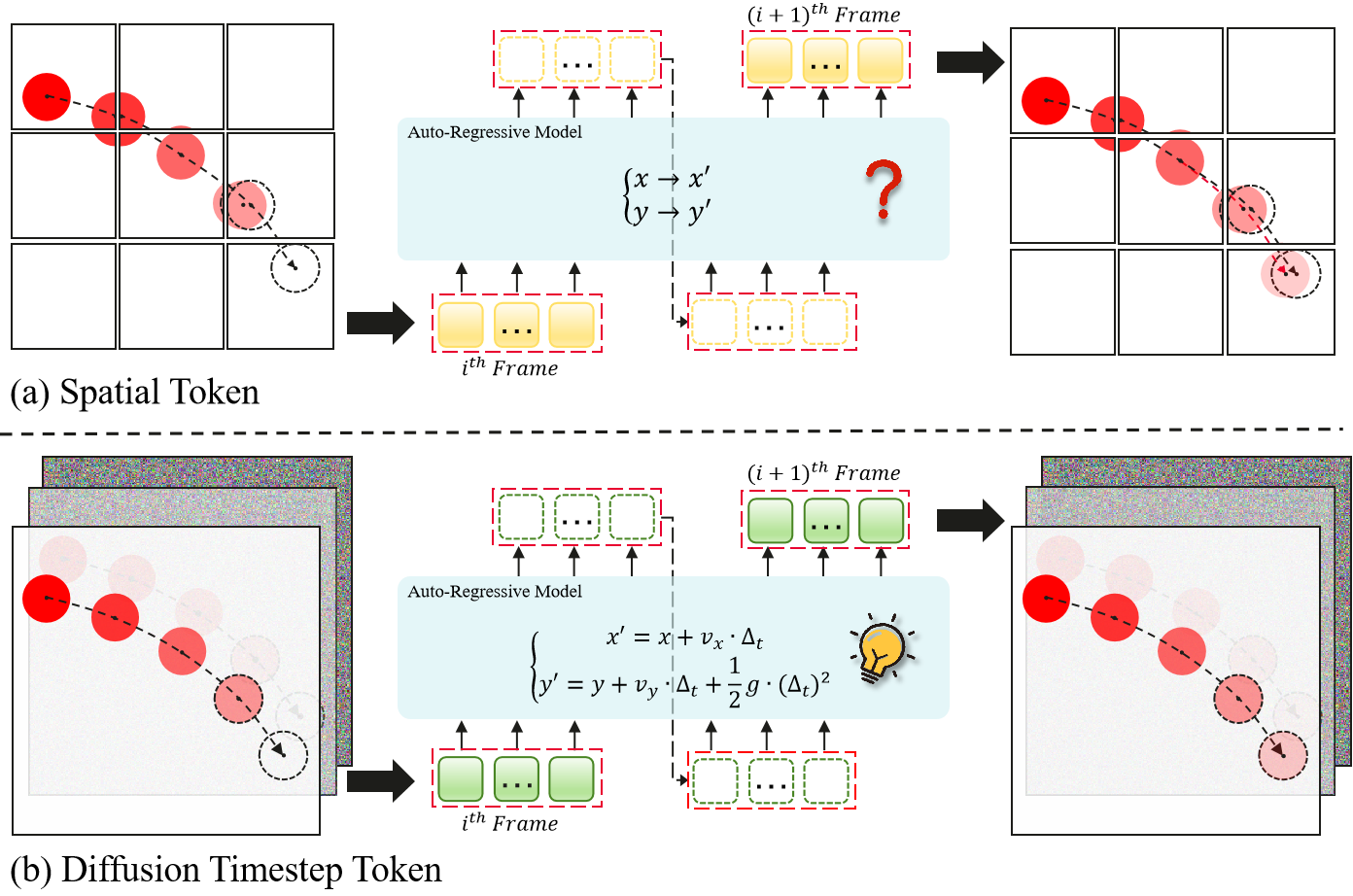}
\vspace{-0.5em}
\caption{Illustration of autoregressive-based(AR) physical video generation.  After given the first 3 frames as conditions: \textbf{(a)} The AR model based on the spatial token generate the ball that deviates from the correct trajectory in the predicted video, which indicates it has not reasoned correctly about the physical laws;  \textbf{(b)} the AR model based on our diffusion timestep token correctly generates a video that conforms to the physical laws.}
\vspace{-1.5em}
\label{fig:intro}
\end{figure}

To achieve this, the first step is to tokenize the video. Most visual tokenizers (\eg, VQGAN~\cite{esser2021taming}) generate spatial tokens from image patches and arrange them in spatial order (\eg, row by row in raster scan order), lacking the hierarchical dependencies required for linguistic-like reasoning. The properties of spatial tokens contradict natural language (see Section~\ref{4.3}) and are therefore, incompatible with the reasoning mechanism of LLMs. As shown in  Figure~\ref{fig:intro}, when using spatial tokens
, the LLM is unable to predict ball trajectories on out-of-distribution data.

Therefore, we propose Diffusion Timestep Tokenizers (DDT) to address the limitations of spatial tokens by learning \textit{recursive structured tokens}. In a Diffusion Model (DM), Gaussian noise is incrementally added at each timestep $t$, gradually erasing visual attributes. Inspired by this, we construct an \textit{expanding sequence} of discrete tokens. Given a noise-free image $\mathbf{x}_0$, we generate a timestep-specific token sequence $f_t(\mathbf{x}_0)$ via an encoder $f$, starting with $f_0(\mathbf{x}_0) = \emptyset$. At $t$, the noisy image $\mathbf{x}_t$ is reconstructed by the DM using $f_t(\mathbf{x}_0)$, ensuring tokens compensate for lost attributes. The token sequence \textit{recursively expands} as $f_{t+1}(\mathbf{x}_0) = \left( f_t(\mathbf{x}_0), \textrm{V}_{t+1} \right)$, where $\textrm{V}_{t+1}$ represents the additional loss in $\mathbf{x}_{t+1}$. At $t = T$, when $\mathbf{x}_T$ is pure noise, the final sequence $f_T(\mathbf{x}_0)$ encodes all visual attributes.
DDT offers the advantage of enabling more effective reasoning by LLMs, bridging the gap between the symbolic nature of language and the visual domain.

Based on the linguistically structured DDT tokens, we propose the two-stage framework \textit{Phys-AR} to enable the co-evolution of physical reasoning and visual generation. In the first stage, supervised fine-tuning (SFT) is used for knowledge transfer of the symbolic system, which expands the DDT tokens into the vocabulary space of the LLMs. This process is similar to the foreign language learning stage in language acquisition, which enables the model to understand the grammatical constraints between symbols, but the distribution characteristics of the training data still limit the inference ability of the model~\cite{chu2025sft}.

The second phase focuses on the derivation of first principles equations within the symbolic system. We reconstruct the video generation as a Markovian decision process at the token level and perform reinforcement learning optimization based on the GRPO~\cite{shao2024deepseekmath} algorithm. Specifically, the reward function of motion parameters is designed to guide the model to explore the essential expression of physical laws so that the LLMs can iteratively revise the inference path through policy gradient updates. This dynamic optimization mechanism breaks through the limitation of pattern solidification in supervised learning and finally realizes the emergent derivation from symbolic operations to physical laws. Our contributions are three-fold:

\begin{itemize}
    \item We propose the Diffusion Timestep Tokenizer, which learns discrete, recursive visual tokens that can be used for symbolic reasoning in LLMs.

    \item We propose an autoregressive-based video generation framework \textit{Phys-AR} that incorporates a symbolic reasoning process into the generation process, thus maintaining the physical correctness of the generated videos.

    \item We demonstrate in three underlying physical motions that PhysAR can reason about physical law and generate physically law-consistent videos.
\end{itemize}

\section{Related Work}
\label{sec:releated}

\subsection{Physical Video Generation}
The integration of physical reasoning into generative models has become an active area of research, aiming to bridge the gap between virtual and physical realities.  
Physics-aware generation methods can be broadly categorized into two approaches: explicit simulation and implicit learning.
Explicit simulation involves directly incorporating physical simulation models into the generation process.   For example, PhysGaussian~\cite{xie2024physgaussian} leverages the Material Point Method (MPM) to simulate physical behaviors of Gaussian splatting, enabling realistic deformation and motion in generated content.   
PhyT2V~\cite{xue2024phyt2v}, GPT4Motion~\cite{lv2024gpt4motion} and PhysGen~\cite{liu2024physgen} leverage the guidance of large language models (LLMs) to generate motion plans or parameters, which are subsequently employed in the rendering of videos.

In contrast, implicit learning relies on models to infer physical properties from data without explicit simulation.  
Models like CogVideoX~\cite{yang2024cogvideox}, VideoCrafter~\cite{chen2024videocrafter2}, and Sora~\cite{sora} show the ability to generate high-quality videos through large-scale pretraining and optimization. 
But~\cite{motamed2025generative} highlights visual realism does not necessarily imply physical understanding.
Recently, some researchers have tried to incorporate latent physical knowledge into generative models. 
~\cite{winterbottom2024power} find that models trained solely on next-frame prediction could predict the values of physical constants (such as gravity) without direct regression task training.
~\cite{cao2024teaching} uses Masked Autoencoders (MAE) to extract latent physical knowledge from data and then integrate this knowledge into the video diffusion models using quaternion networks.
~\cite{kang2024far} conducted empirical studies showing the limitations of these models in understanding physical principles and the generalization to unseen physical conditions. Therefore, in this paper, we explore autoregressive video generation, attempting to integrate symbolic reasoning into the video generation process to maintain physical principles.

\label{sec:methods}
\begin{figure*}[t]
\centering
\includegraphics[width=\textwidth]{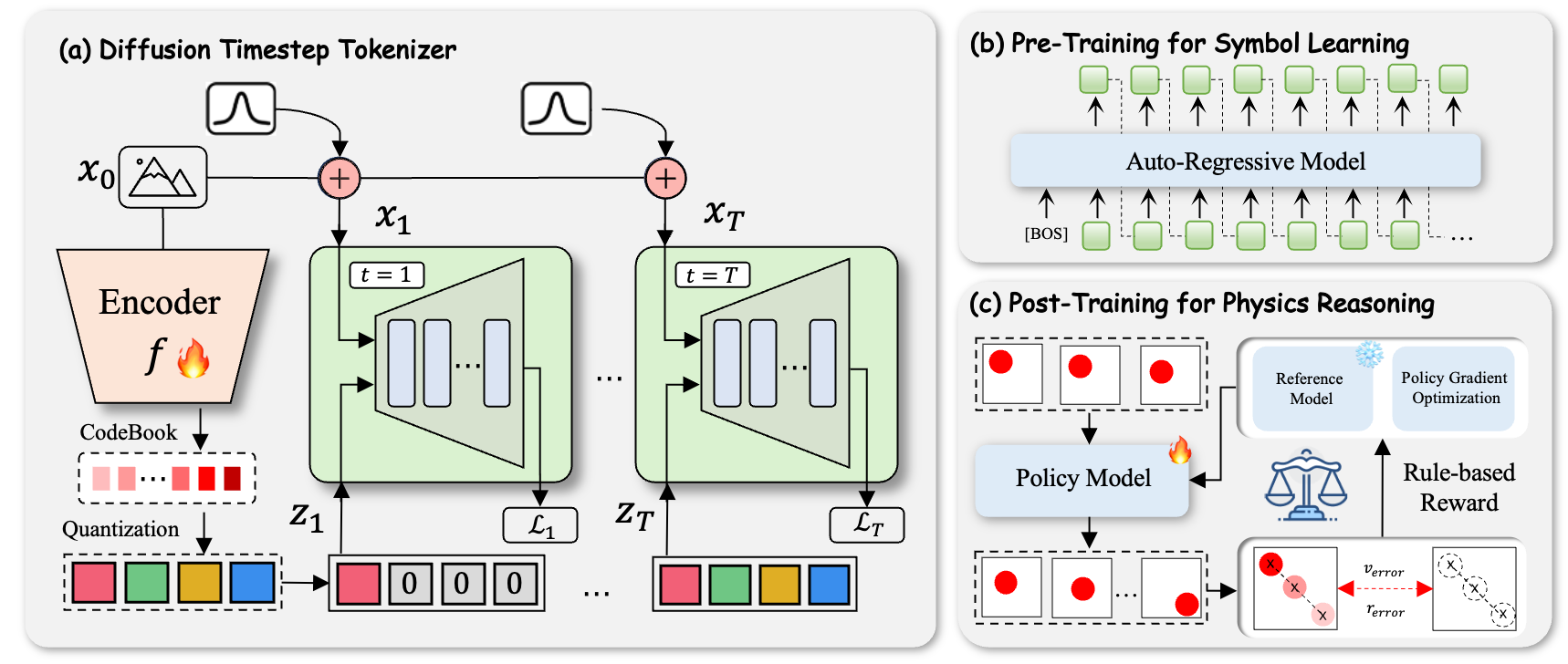}
\vspace{-2em}
\caption{The overview of our methods. \textit{\textbf{(a):}} The architecture of diffusion timestep tokenizer encodes an image to a recursive sequence of discrete tokens. \textit{\textbf{(b):}} An Auto-Regressive architecture which learns new image tokens based on next token prediction. \textit{\textbf{(c):}} Reinforcement learning infers physical laws through reward signals based on physical variables.}
\vspace{-1em}
\label{fig:overview}
\end{figure*}

\subsection{Reinforcement Learning in Visual Generation}
Recent breakthroughs in large-scale inference models, such as OpenAI’s o1~\cite{jaech2024openai} and DeepSeek-R1~\cite{guo2025deepseek}, have greatly enhanced the capabilities of language models in various tasks, including reasoning and decision-making. Building on the success of reinforcement learning with human feedback (RLHF) in large language models (LLMs), researchers have begun exploring the application of RLHF to visual generation tasks.
In the field of image generation, several approaches have introduced feedback-based learning to align diffusion models with human preferences. DDPO~\cite{black2023training} employs reinforcement learning to guide diffusion models using supervision from additional reward models. In contrast,~\cite{wu2023human, wu2023human-v2, xu2023imagereward} uses a reward model trained on preference data to filter preferred outcomes, enabling feedback learning through supervised fine-tuning.
For video generation, recent studies~\cite{yuan2025self, liu2024videodpo, zhang2024onlinevpo} focus on aligning diffusion-based or flow-based video generation models by annotating videos generated from earlier open-source models. Our work differs in two key ways: first, we explore reinforcement learning for autoregressive video generation, and second, we prioritize reasoning about the physical laws that govern the video generation process, rather than aligning with human preferences.

\section{Methods}
\subsection{Formulation of Physics Video Generation}
First, we provide a definition of physics simulation and physics generation.
Physics simulation is the process of evolving the input State condition $S$ into the output State $S'$ derived from first-principle equations as 
$F_{\theta}(S)\to S'$, where the observations $S$, $S'$ can be from different simulation time steps. For instance, in free-fall motion, $F_{\theta}$ can be expressed  as $ \frac{1}{2}gt^{2} $.
Physics generation is the process of as
$G(I)\to I'$, which creating the next frame $I'$ from the input frame $I$ using the generative model $G$. 

In this work, we aim to generate and simulate simultaneously as follows:
\begin{equation}
\small
    G_{F_{\theta}}(I)\to I',
    \label{eq:1}
\end{equation}
Next, we first describe how to convert video frames $I$ to state conditions $S$ in Section~\ref{3.2}, present how the generative model $G$ masters the state conditions $S$ in Section~\ref{3.3}, and finally illustrate how to derive the physical principles $F_{\theta}$ in Section~\ref{3.4} and generate the final Physical Video.

\subsection{Symbolising Video with DDT Tokens}
\label{3.2}
Let $I$ denote a video frame,
our goal is to use an image tokenizer that encodes an image to a recursive sequence of discrete tokens---to be like state condition $S$---and decode the tokens back to the image---to render the tokens of $I'$ predicted by the LLM during inference. 
As shown in Figure~\ref{fig:overview} (a), the model consists of an encoder, a quantizer, and a diffusion model $d$ as a decoder. We provide details on each parts below.

\noindent\textbf{Encoder}. 
Given an video frame $I$, we first use an off-the-shelf pre-trained variational autoencoder (VAE) model to extract its latent features $\mathcal{V}=\texttt{VAE}_{enc}(I)$, which are then fed into the encoder $\texttt{Enc}$ with $T$ 1D learnable query tokens $\mathcal{Q} \in R^{T\times d}$ (we set $T$=16). 
The Q-former encoder has two independent transformers to separate the visual latent $\mathcal{V}$ and the query tokens $\mathcal{Q}$. But it joins the sequences of the two modalities for a unified-direction attention operation, such that both representations can work in their own space yet take the other one into account. 
In the encoder output, we only retain the query latent $\texttt{Enc}(\mathcal{Q}, \mathcal{V})[:T]$ as the image’s latent representation, thereby enabling a more compact latent representation of 1D sequence $Z_{1D}=\{z_0,...,z_K\}$.

\noindent\textbf{Quantizer}. 
The query latents $Z_{1D}$ are then passed into a vector quantizer, which are tokenized to a sequence of discrete timestep codes by looking up a learnable codebook $\mathcal{C}=\{c_n\}_{n=1}^N$ ($N=256$ is the codebook size) according to cosine similarity. 
Note that we use an EMA-variant of VQ to improve training stability, \ie, updating $\mathcal{C}$ by gradually moving each entry towards the cluster center of its matched encoder outputs.
Inspired from~\cite{zeghidour2021soundstream}, we monitor dead entries in $\mathcal{C}$ at each training step (\ie, rarely matched with any $z_i$) and set them to random $z_i$ in a training batch. 

\noindent\textbf{Decoder}. 
The decoder is similar to MMDiT~\cite{esser2024scaling}. Building upon the DiT architecture, it leverages two independent transformers for the visual noise sequence and timestep tokens $\mathbf{Z}_{1D}$ (serving as the condition), respectively, and combines both sequences for the attention operation. After the final DiT block, we rearrange the denoised output of the image stream back into its original spatial layout and use the pre-trained VAE decoder to decode it into images.

\noindent\textbf{Training}. 
Recall that we aim to learn a recursive sequence in the form $f_{t}(\mathbf{x}_0)=\left(f_{t-1}(\mathbf{x}_0\right), \textrm{V}_{t})$ at each $t$, such that $f_{t}(\mathbf{x}_0)$ makes up for the attribute loss in the noisy image $\mathbf{x}_t$. Hence we use an expanding set of tokens $({\mathrm{V}}_1,\ldots,{\mathrm{V}}_t)$ as the input to the decoder $d$, and train everything end-to-end with the reconstruction loss:
\begin{equation}
\small
    \mathcal{L} = \mathop{\mathbb{E}}_{t, \mathbf{x}_0, \epsilon} \; \underbrace{\left[ \lVert  d \left( t \epsilon + (1-t) \mathbf{x}_0 , t, ({\mathrm{V}}_1,\ldots,{\mathrm{V}}_t)\right) - \mathbf{x}_0 \rVert^2 \right]}_\text{$\mathcal{L}_t$},
    \label{eq:6}
\end{equation}
where we follow Rectified Flow~\cite{liu2022flow} to sample the noisy $\mathbf{x}_t = t \epsilon + (1-t) \mathbf{x}_0$ with the Gaussian noise $\epsilon$. We also use a standard commitment loss $\sum_{i=1}^T \lVert \hat{\mathrm{V}}_i - \mathrm{V}_i \rVert^2$ to regularize the encoder output (continuous vector) to be similar to the matched quantized vector.

\subsection{Pre-Training for Symbol Learning}
\label{3.3}
With diffusion timesteps tokenizer, we can transform visual content into 1D discrete token sequences which have language-like structure. In this stage, we aim to train a LLM to learn these tokens in the same way it would learn foreign language.

We begin by extending the vocabulary of the pre-trained LLM by introducing $\mathcal{C}=256$ additional visual codes. The LLM then adopts the general Language Modeling (LM) objective, which is to maximize the likelihood of frame sequences in an autoregressive manner. Specifically, we aim to optimize the model’s ability to predict the next token in the sequence, conditioned on the previous tokens. The objective function for this task is given by:
\begin{equation}
p(y) = \sum_{y \in \mathcal{C}} \sum_{i=1}^{S} \log G(y_i | y_{< i}),
\end{equation}
where $y$ represents the visual token and $S$ indicate as sequence length. Since the frames have been transformed into discrete token IDs, we treat these IDs as the ``words'' in the sequence and use the standard language modeling objective to predict the next token. This approach allows the model to learn the structure and flow of the visual content by treating video frames as a sequence of symbolic elements.

To supervise the learning process, we use the cross-entropy loss function, which measures the difference between the predicted token probabilities and the true token labels. This enables us to guide the model towards accurate predictions and reinforce its understanding of the relationships between visual tokens. Throughout the training process, all parameters of the LLM are meticulously fine-tuned, ensuring that the model can effectively integrate the newly added visual tokens.
By the end of this pre-training stage, the LLM will have learned to process video frames in a symbolic manner, allowing it to generate sequences of visual tokens that represent the video content.

\begin{figure*}[t]
\centering
\includegraphics[width=\textwidth]{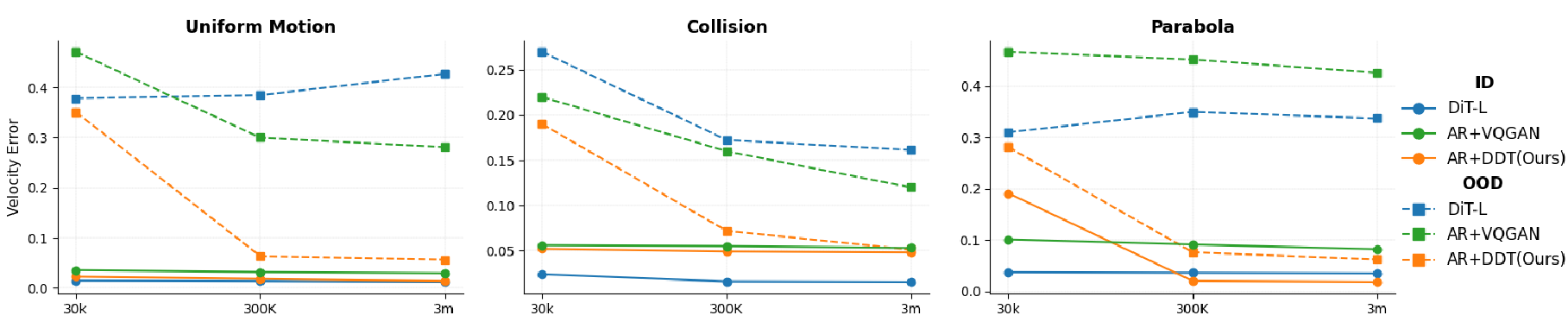}
\vspace{-1em}
\caption{Comparison of the velocity errors of the balls using different methods for generating in-distribution and out-of-distribution videos, given the first three frames as input. The prediction error of the DDT-based AR model on out-of-distribution data is at the $10^{-2}$ magnitude, while the errors of the DiT model and the spatial-based AR model are at the $10^{-1}$ magnitude.}
\vspace{-1em}
\label{fig3}
\end{figure*}

\subsection{Post-Training for Physics Reasoning}
\label{3.4}
Through the above process, we use a diffusion tokenizer to transform candidate video frames into representations that are close to the state condition, \textit{i.e.}, $\text{DDT}(I) \approx S$. With pre-training, we obtained a generative model $G$ that can generate the next frame's token, \textit{i.e.}, $G(\text{DDT}(I)) \rightarrow \text{DDT}(I')$. Now, we move to the final step of deriving the physical law $F_{\theta}$ in the generative model by symbolic reasoning.

\textbf{Reinforcement Learning with Rule-based Rewards.}
Although pre-trained language models trained based on the next token prediction paradigm can grasp the surface laws of the symbolic system, there are still significant limitations in their cognitive abilities---as humans' mastery of the Arabic numeral symbolic system is not equivalent to having mathematical reasoning abilities. In order to break through this cognitive bottleneck, inspired by recent reasoning models~\cite{guo2025deepseek}, we propose to adopt the reinforcement fine-tuning, which guides the model to independently discover the implicit physical laws in the video data.

We designed the reward function based on two fundamental physical quantities that describe the motion: velocity and mass. 
For velocity, it not only determines the speed and direction of an object’s motion but also relates closely to physical concepts such as energy, momentum, time, and space, which are essential for understanding and analyzing physical motion. Following~\cite{kang2024far}, the velocity error is computed by examining the trajectory of the center of the circle between adjacent frames, as expressed by the equation:
\begin{equation}
\text{v}_{error} = \frac{1}{N|T|} \sum_{i=1}^{N} \sum_{t \in T} \left| \mathbf{v}_t^i - \hat{\mathbf{v}}_t^i \right|, 
\end{equation}
where $\mathbf{v}_t^i$ represents the computed velocity at time $t$, $\hat{\mathbf{v}}_t^i$ denotes the ground-truth velocity from the simulator, $N$ is the number of balls, and $|T|$ is the number of valid frames.

Then, we introduce a reward function to convert the error to the excitation signal:
\begin{equation}
R_{\text{vel}}=\alpha e^{-k\text{v}_{error}}
\end{equation}
where $\alpha$ controls the maximum reward and $k$ determines the reward decay rate, allowing for a differentiated response to various error intervals. This nonlinear mapping strategy creates a reward space with a steep gradient in the low error region ($\text{v}_{error} \rightarrow 0$), enhancing the model’s sensitivity to accurate predictions.

For mass, we assume that the ball’s density is uniform, meaning that the radius represents the mass. The radius is calculated by detecting the mask of the region where the ball is located. The reward function for mass is given by:
\begin{equation}
R_{\text{mass}}=1-\text{r}_{error}
\end{equation}
where $\text{r}_{error}$ denotes the radius error. Finally, the overall reward function is defined as:
\begin{equation}
R = R_{\text{vel}}+R_{\text{mass}}
\end{equation}
This combined reward function integrates both velocity and mass errors, providing comprehensive feedback to optimize the model’s prediction.

\textbf{Training.} The RL training process follows the SFT phase, enhancing the model’s ability to handle complex reasoning tasks.  
Specifically, we use the Group Relative Policy Optimization (GRPO)~\cite{shao2024deepseekmath}  framework.
Different from reinforcement learning algorithms such as PPO~\cite{schulman2017proximal} that require a critic model to evaluate policy performance, GRPO compares groups of candidate responses directly, eliminating the need for
an additional critic model. 
Given the input condition frames $c$, GRPO first generates $g$ distinct predictions $\{p_{1},p_{2},...,p_{g}\}$ from the current policy $\pi_{\theta_{old}}$. Then, GRPO takes actions based on these predictions and denotes the obtained rewards as $\{r_{1},r_{2},...,r_{g}\}$. By computing their mean and standard deviation for normalization, GRPO determines the relative quality of these responses: 
\begin{equation}
A_i = \frac{r_i - \text{mean}(\{r_1, \dots, r_g\})}{\text{std}(\{r_1, \dots, r_g\})}, 
\end{equation}
where \( A_i \) represents the relative quality of the \( i \)-th answer. GRPO encourages the model to favor better answers with a high reward value within the group. GRPO optimizes the following objectives:
\begin{align}
&\max_{\pi_\theta} \mathbb{E}_{p \sim \pi_\theta(c)} \left[ R(c, p) \right] \\
&\quad = \left[ R(c, p) \vphantom{\text{KL}} \right.\left. - \beta \, \text{KL}[\pi_\theta(p|c) \parallel \pi_{\text{old}}(p|c)] \right]
\end{align}
where $ R $ is the reward function, and $ \beta $ is the hyperparameter to control the KL-divergence. The verifiable reward function $ R $ takes the condition and predicted frames $ (c, p) $ as inputs and checks if the ground-truth answer remains the same as the prediction $p$.

After this, the model receives rewards or penalties based on adherence to the rules. By constructing a dynamic exploration-reward feedback closed loop of physical laws, the model gradually establishes the ability to characterise the underlying laws, such as Newton's laws of mechanics.

\section{Experiment}
\label{sec:experiment}

\begin{figure*}[t]
\centering
\includegraphics[width=\textwidth]{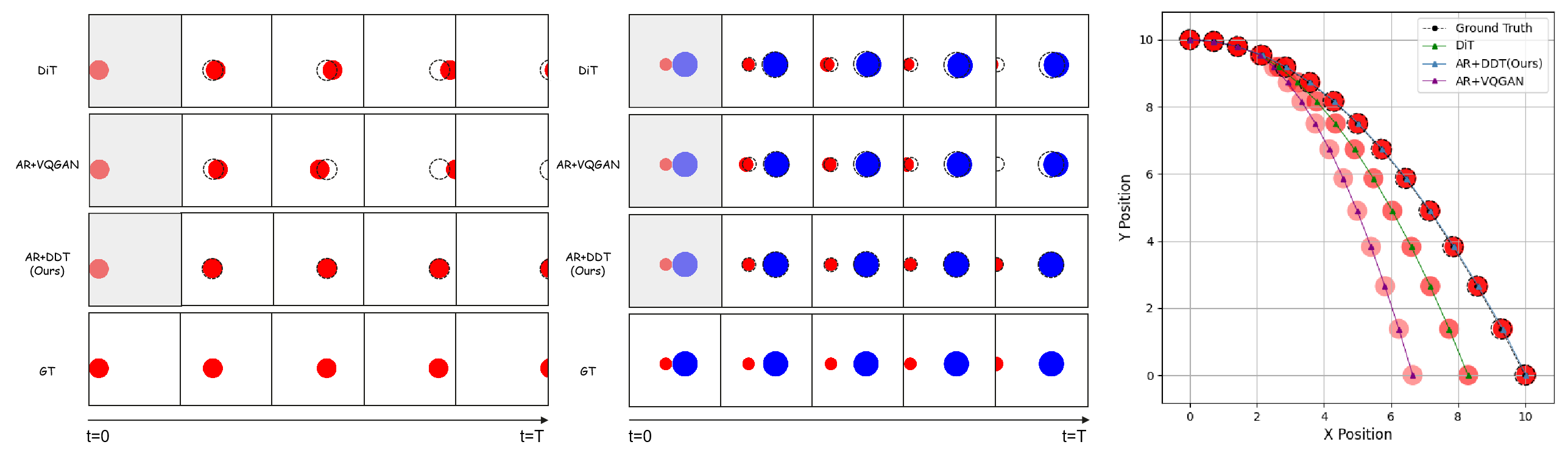}
\vspace{-2em}
\caption{Comparison of the generated video results for the 3 physical motions, the dotted line indicates the correct motion trajectory, and the arrow indicates the progression of time. For the parabolic, we show the predicted video frames superimposed for ease of comparison. }
\vspace{-1em}
\label{fig:case}
\end{figure*}

\subsection{Experiment Settings}

\noindent\textbf{Dataset.} 
We use the videos of 3 physical motions (uniform motion, parabola, collision) from PhyWorld~\cite{kang2024far}, which has a world as a $10 \times 10$ grid, with a timestep of 0.1 seconds, resulting in a total time span of 3.2 seconds (32 frames).
We take the first 3 frames of the video as input and predict the subsequent 29 frames.
In each video, the ball is set to a different radius (which determines the mass) and initial velocity. For all scenarios, we choose the data with radius $r \in [0.8, 1.4]$ and velocity $v \in [1.5, 3.5]$ for pertaining.  In post-training, we choose the data with the radius $r \in [0.7, 0.8] \cup [1.4,1.5]$ and velocity $v \in [1.0,1.5] \cup [3.5,4.0]$.
For evaluation, out-of-distribution (OOD) ranges are defined as $r \in [0.3, 0.6] \cup [1.5, 2.0]$ and $v \in [0, 0.8] \cup [4.5, 6.0]$.

\noindent\textbf{Tokenizer Training.}
Our diffusion timestep tokenizer encodes each image into 16 tokens with a vocabulary of size 256. Besides, each training image is resized to a size of $128\times 128$. In order to maintain a fair comparison, the spatial tokenizer follows the same settings and achieves the same performance in terms of reconstruction quality; see the Appendix for more details.

\noindent\textbf{Pre-Training.}
We initialize our LLM from a pre-trained LLM, specifically using the Llama3.1-8B~\cite{grattafiori2024llama} model. The training process is conducted with a batch of 768 and a learning rate of $ 1\times 10^{-4}$. During inference, when generating the image, we mask the logits of the text words before sampling the image tokens.

\noindent\textbf{Post-Training.}
The parameters of LLM are fully fine-tuned in reinforcement learning. We set a learning rate of $ 3\times 10^{-6}$ with a batch of 28.
Besides, we also mask the text words and set $topk=50$ and $topp=0.95$ for image token sampling.

\subsection{Main Results}
\begin{table}[]
\centering
\scalebox{0.64}{
\begin{tabular}{lccccccc}
\toprule
Methods & \multicolumn{2}{c}{Uniform}                 & \multicolumn{2}{c}{Parabola}                & \multicolumn{2}{c}{Collision}               \\ \cline{2-3}  \cline{4-5} \cline{6-7}
                         &$\text{IID}_{error}$ & $\text{OOD}_{error}$ &$\text{IID}_{error}$ & $\text{OOD}_{error}$  &$\text{IID}_{error}$ & $\text{OOD}_{error}$ \\
                         \hline
DiT-S                    &0.015 &0.288          &0.052         &0.311       &0.023          &0.153                  \\
DiT-B                   &0.014         &0.358          &0.036          &0.287           &0.018            &0.211                    \\
DiT-L                    &0.012            &0.427           &0.035          &0.337          &0.015             &0.161                    \\
\hline
Qwen-1.5B                  &0.025            &0.271          &0.023          &0.399          &0.056            &0.185                    \\
Qwen-7B                     &0.016            &0.071          &0.017          &0.068          &0.043            &0.047               \\
Llama3.1-8B                  &0.014         &0.057          &0.018          &0.062          &0.048            &0.051              \\
\bottomrule
\end{tabular}
}
\vspace{-0.5em}
\caption{Comparison of different families of model architectures and the effect of the number of model parameters on the velocity error of generated video.}
\vspace{-1em}
\label{tab:modelsize}
\end{table}

We compared the diffusion-based model (DiT), the autoregressive model based on spatial tokens (AR+VQGAN), and the autoregressive model based on diffusion timestep tokens (AR+DDT). The experimental results, shown in Figure~\ref{fig3} and~\ref{fig:case}, indicate that although DiT and AR+VQGAN achieved small errors on in-distribution data, their generalization ability on out-of-distribution data remains limited. This suggests that these models rely more on memorizing the training data rather than truly understanding and mastering physical laws. In contrast, AR+DDT demonstrated significantly better performance on out-of-distribution data, showing good generalization ability. This implies that AR+DDT is capable of inferring corresponding physical conditions based on the input video frame conditions and generating physically consistent videos. Furthermore, we conducted ablation experiments to analyze the performance differences of the models under various settings, focusing on the training data scale and language model parameters.

\noindent\textbf{Ablation Study on Training Data.} As shown in Figure~\ref{fig3}, we trained the models on datasets of three different sizes. For in-distribution data, all models showed decreasing errors with the increasing training data size. However, for out-of-distribution data, DiT did not show significant error reduction as the data size increased, and it still performed poorly with larger data, indicating that the model’s generalization ability is limited. The autoregressive models performed with larger errors on smaller datasets (30k), especially in out-of-distribution data, which can be attributed to insufficient pretraining of image tokens, failing to capture the global features of the data effectively. As the training data size increased and pretraining conditions were met, the autoregressive models gradually showed improved performance, indicating that these models are less sensitive to data size once pretraining requirements are satisfied.

\noindent\textbf{Ablation Study on Generative Model Architecture.} As shown in Table~\ref{tab:modelsize}, we compared models with different parameter sizes. For the diffusion-based method, increasing the model parameters did not significantly reduce the prediction error. For instance, in the uniform motion task, DiT-S had the lowest error, while DiT-L had the highest, suggesting that increasing model parameters does not necessarily lead to better performance in diffusion models. In autoregressive models, when the parameter size is small (such as Qwen-1.5B~\cite{yang2024qwen2}), the model lacks sufficient reasoning power, resulting in higher out-of-distribution errors. This indicates that autoregressive models with smaller parameters do not have enough reasoning capacity to effectively capture complex physical laws. However, as the model size increases, the effect of different large language model architectures on prediction errors becomes minimal, suggesting that larger language model architectures have stronger reasoning capabilities, enabling them to better handle complex video generation tasks.

\subsection{Why are DDT tokens like language?}
\label{4.3}

\noindent\textbf{DDT Tokens are Recursive.} To directly show that DDT tokens are recursive and decouple visual attributes at various granularities, we first conduct counterfactual interpolation with both DDT tokens and VQGAN tokens. 
Specifically, for the token sequence derived from an image, we replace a subset of tokens with those from another image while keeping the remaining tokens fixed, with the resulting sequence fed into the decoder for image generation.

As shown in Figure~\ref{fig:interpolation}, for VQGAN tokens, counterfactual interpolation actually performs CutMix~\cite{yun2019cutmix}, where regions from two images are concatenated. 
In contrast, DDT tokens, with their disentangled representation, ensure that only the attributes captured by the substituted tokens change in the generated counterfactuals. This behavior aligns with our expectations for the DDT tokens design.

In contrast, DDT tokens can untangle the attributes that represent the image, \textit{i.e.}, certain tokens represent the radius of the ball and certain tokens represent the position of the ball. 
This ensures that only the attributes captured by the replacement tokens change in the generated counterfactuals, \textit{e.g.}, the radius and position of the ball that the DDT token changes when interpolated.

\begin{figure}[t]
\centering
\includegraphics[width=0.43\textwidth]{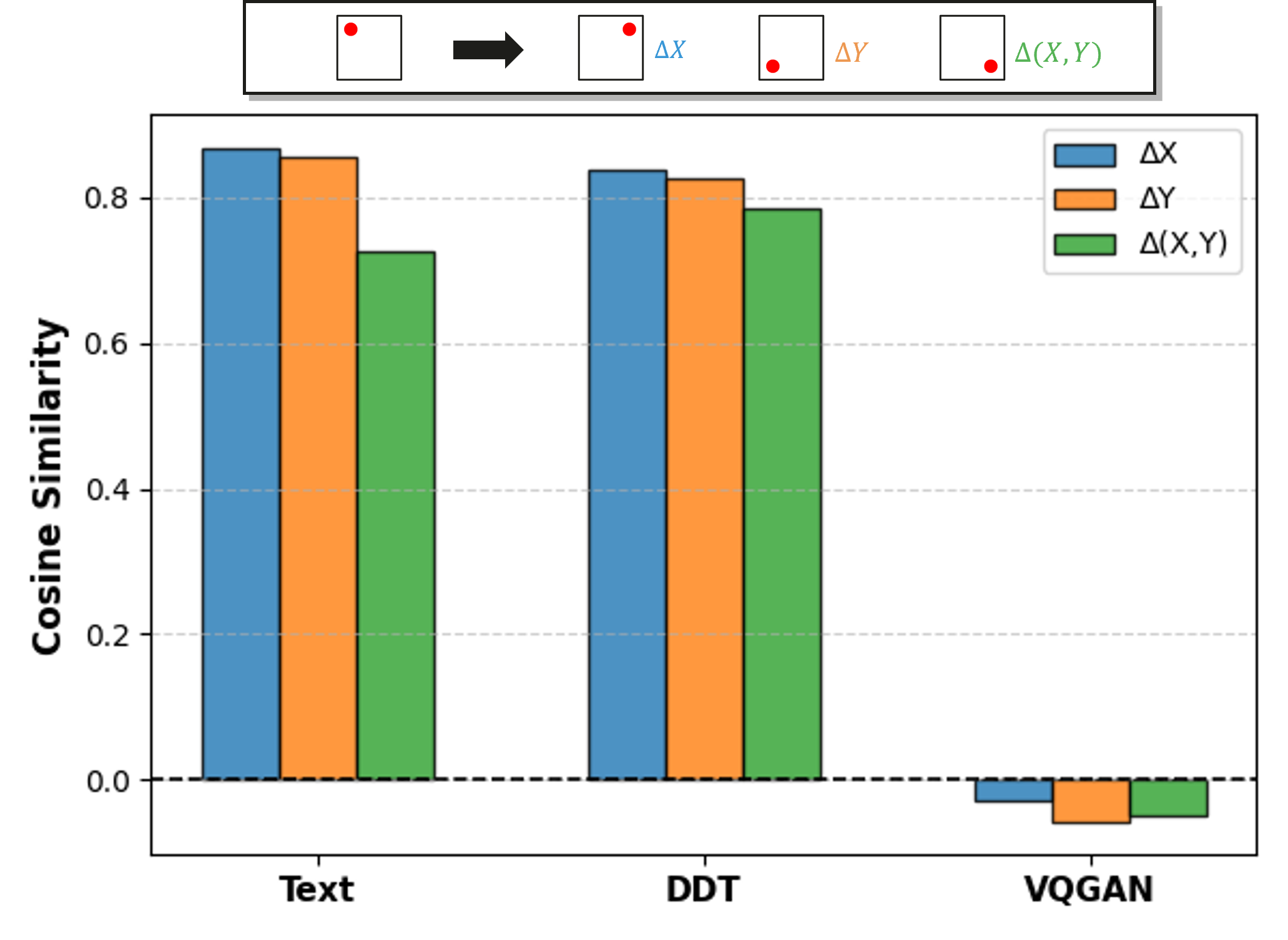}
\vspace{-1em}
\caption{Comparison of the response of different token embedding to image changes. The response of DDT token is basically the same as that of text token, while VQGAN does not behave the same as text token due to the symmetry of 2D image space.}
\vspace{-2em}
\label{fig:xy_change}
\end{figure}

\noindent\textbf{DDT token is consistent with language in response to image attribute changes.}
The physical state of the ball in each frame can be described by a pair of coordinates (representing the $x$ and $y$ positions). We randomly modify these coordinates on the original image, such as changing the coordinates from (0.5, 0.5) to (1.2, 0.5). We randomly generated $5k$ such combinations and compared whether the changes in the two image token embeddings align with those in the text token embeddings in the LLM.

As shown in Figure~\ref{fig:xy_change}, the results show that the changes in DDT token embeddings are closely similar to those in the text token embeddings. However, the cosine similarity of the spatial token embeddings is negative.
The reason is the modified image exhibits a certain degree of spatial symmetry with the original image. This symmetry leads to a semantic "antonym", which does not exist in the original language text. This "antonym" nature is injected into the semantic of the spatial tokens, thereby disrupting the ability to reason about physical laws based on spatial tokens.

On the other hand, our diffusion timestep tokenizer is trained by compensating for the missing attributes at different timesteps of the diffusion model and is not disturbed by spatial ordering. It exhibits semantic changes similar to those of text tokens when faced with positional changes. This allows LLMs to read DDT tokens as a new type of numerical symbol (\textit{e.g.}, ``7'' $\to$ \text{``VII''} ).

\begin{figure}[t]
\centering
\includegraphics[width=0.45\textwidth]{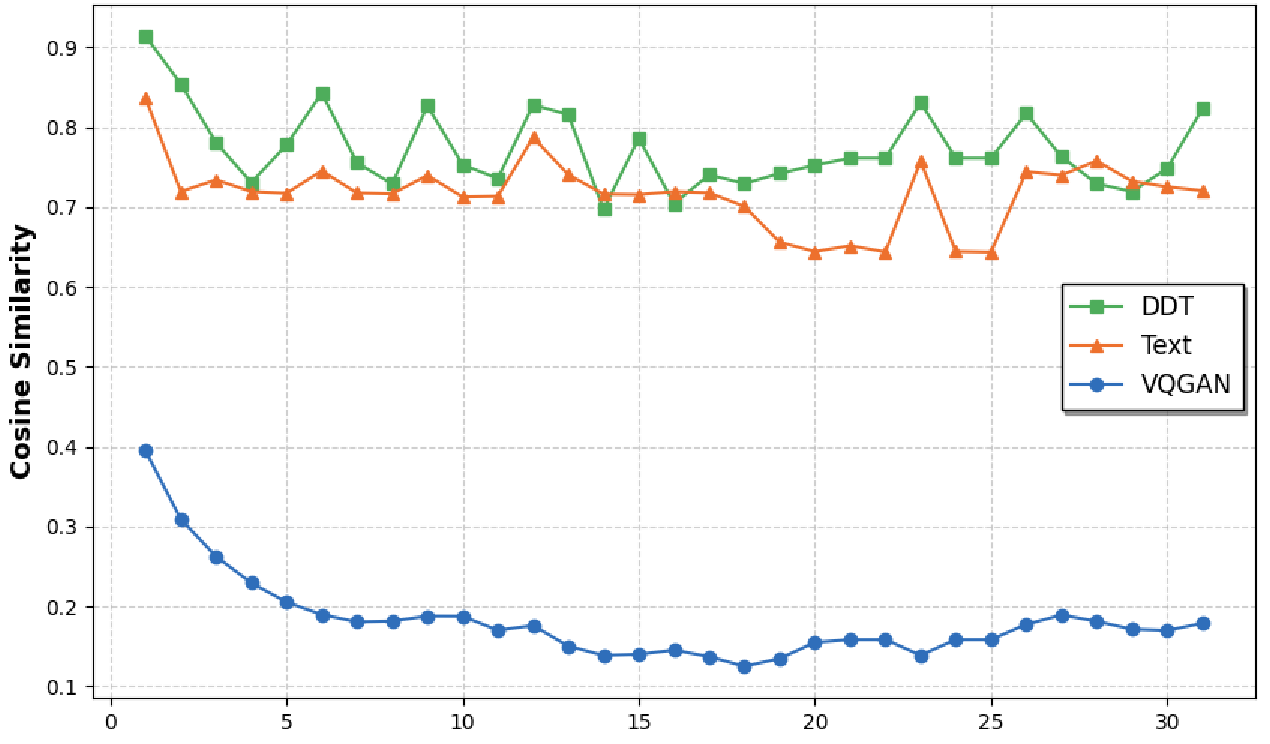}
\vspace{-0.5em}
\caption{Comparison of token embedding similarity between each frame and the next frame for different tokenizers. Motion parameters of adjacent frames digitally represented by coordinates have higher and more uniform feature similarity in text tokenizer. The variation of DDT tokens is consistent with the text tokens.}
\vspace{-1em}
\label{fig:frame_chaneg}
\end{figure}

\begin{figure}[t]
\centering
\includegraphics[width=0.48\textwidth]{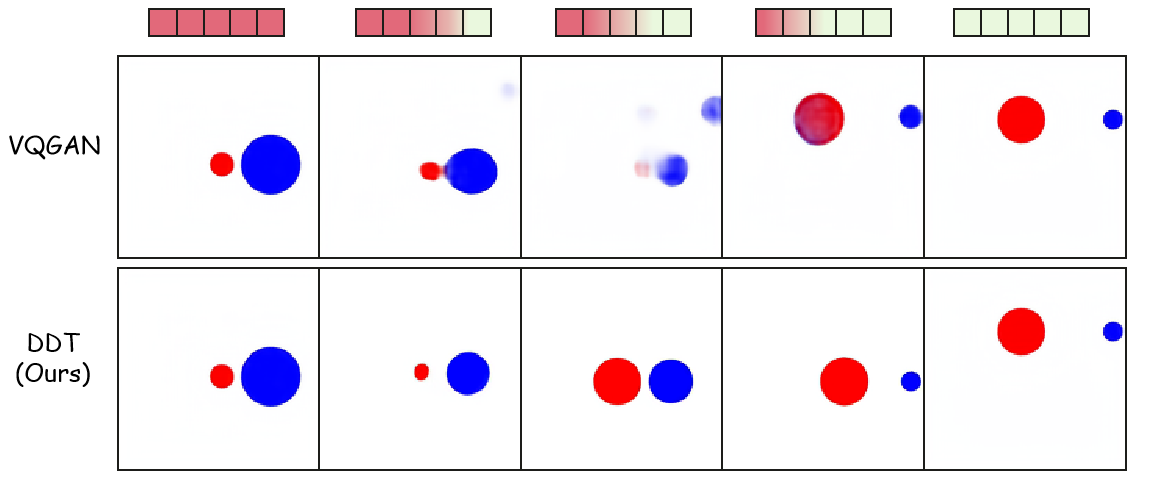}
\vspace{-1em}
\caption{Results of Counterfactual Interpolation with DDT tokens and VQGAN tokens. With token substitution, the VQGAN tokens behave like pixel interpolation, while the DDT tokens exhibit an exchange of image attributes (\textit{e.g.}, radius, position) }
\vspace{-1.5em}
\label{fig:interpolation}
\end{figure}

\begin{figure*}[t]
\centering
\includegraphics[width=\textwidth]{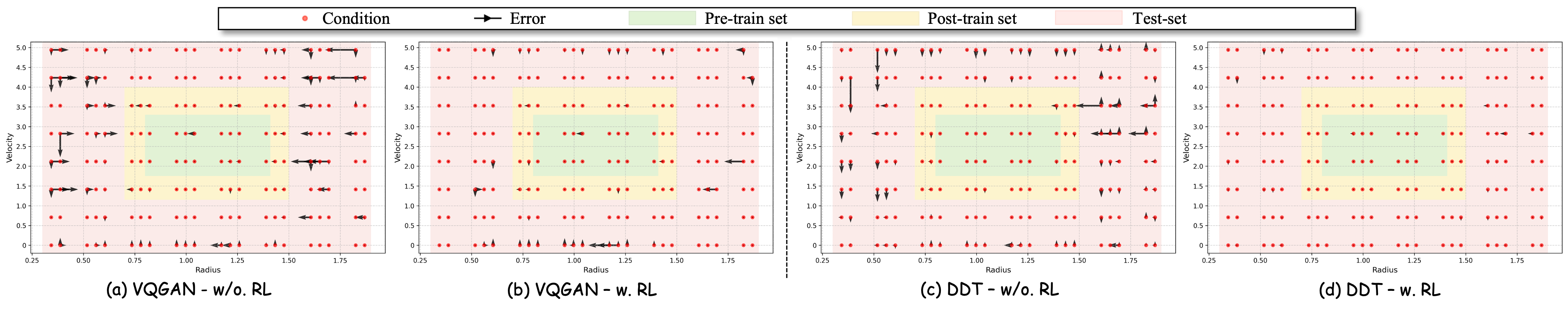}
\vspace{-1.5em}
\caption{Impact of Reinforcement Learning on Different Tokenizers: We visualized the changes in error for both tokenizers before and after reinforcement learning. The horizontal (radius error) and vertical  (velocity error) arrows ($\rightarrow$) represent the shift between the generated video and their initial conditions, \ie, the error. The green area represents the range of the pretraining data, the yellow area represents the range of data used in reinforcement learning, and the red area represents the range of test data.}
\vspace{-1em}
\label{fig:error}
\end{figure*}

\noindent\textbf{DDT token is consistent with language in response to dynamic changes in video.} 
We further investigate the variation trend of 3 types of token embeddings in adjacent video frames. We report statistical results on 1000 randomly selected videos.
As shown in Figure~\ref{fig:frame_chaneg}, our DDT tokens and text tokens are highly similar in terms of changes between adjacent frames. However, the embedding similarity of the spatial token is significantly lower. And the similarity of token embedding for the video frames that are close to the back further reduces because of the acceleration that causes larger pixel changes. This indicates that there is greater semantic variation between adjacent frames represented by spatial tokens, which hinders the accuracy of next-frame prediction and also hinders LLM from generalizing physical laws from video sequences.
On the contrary, the high token embedding similarity between adjacent frames of our DDT token indicates that the semantic change between frames is smoother and the LLMs are easier to reason about.

\subsection{How to reason the physical laws?}

\begin{table}[]
\centering
\scalebox{0.59}{
\begin{tabular}{cl ll ll ll}
\toprule
\multicolumn{2}{c}{\multirow{2}{*}{Methods}} & \multicolumn{2}{c}{Uniform}                 & \multicolumn{2}{c}{Parabola}                & \multicolumn{2}{c}{Collision}               \\ \cline{3-4}  \cline{5-6} \cline{7-8}
                         & &$\text{IID}_{error}$ & $\text{OOD}_{error}$ &$\text{IID}_{error}$ & $\text{OOD}_{error}$  &$\text{IID}_{error}$ & $\text{OOD}_{error}$ \\
                         \hline
\multirow{2}{*}{VQGAN}       & w/o. RL                &0.032            &0.884          &0.090          &0.951          &0.054            &0.322                    \\
& w. RL                   &0.030            &0.281          &0.082          &0.426          &0.053            &0.120                    \\
\hline
\multirow{2}{*}{DDT}         & w/o. RL                    &0.032            &0.763          &0.024          &0.685          &0.050            &0.184                    \\
& w. RL                    &0.014         &0.057          &0.018          &0.062          &0.048            &0.051                    \\
\bottomrule
\end{tabular}
}
\vspace{-0.5em}
\caption{Comparing the performance of reinforcement learning with different tokneizers. Reinforcement learning significantly reduces the prediction error of DDT tokens in OOD videos, while VQGAN-based spatial tokens have limited gains.}
\label{tab:ablation}
\vspace{-1em}
\end{table}

\noindent\textbf{Reinforcement learning can improve generalization.}
We compared the changes in error before and after reinforcement learning for the two tokenizers. As shown in Figure~\ref{fig:error}, for DDT tokens, we observed that when radius was small, the generated ball had almost no error; however, when the radius was large, the generated ball tended to be smaller than expected. This suggests that the model has a bias towards generating smaller objects. In terms of speed, the DDT approach showed a tendency to generate slower motion. After undergoing reinforcement learning (RL) training, the AR model based on DDT tokens significantly improved, achieving smaller errors on all out-of-distribution (OOD) test data. The use of DDT tokens enabled the model to effectively reason about physical properties and adjust its predictions in line with the new data, resulting in improved accuracy across a broader range of scenarios.

In contrast, the AR model based on spatial tokens demonstrated larger errors after pre-training. Although the model showed some improvement after reinforcement learning, the reduction in error was limited, and the model’s performance still degraded as the distribution of the data differed more from the training set. For example, when the speed was 0 or 5.0, the error remained substantial, indicating that the spatial tokens struggled to handle some extreme conditions. This highlights the inefficiency of reinforcement learning with spatial tokens, resulting in the model’s generalization ability was significantly limited.
 
\noindent\textbf{DDT tokens are better suited for reinforcement learning exploration.}
We compare the performance of reinforcement learning using different image tokenizers. As shown in Table~\ref{tab:ablation}, we observe that reinforcement learning on spatial tokens yields inferior results compared to the DDT token-based approach. This can be attributed to the recursive structure inherent in DDT tokens. In reinforcement learning, discovering new combinations of states and actions often involves exploring various paths multiple times. Language tokens, with their recursive structure, offer a natural advantage in this exploration process. For example, in language, an expression like ``20=((((16)+1)+1)+1)'' allows for backtracking during the exploration, which significantly improves the efficiency of the learning process. This recursive property might enable the model to implicitly revisit previous steps and adjust its actions accordingly, optimizing exploration in fewer steps. In contrast, spatial tokens lack this recursive structure and do not allow for backtracking. As a result, spatial tokens require a more extensive search process, leading to higher computational complexity during exploration. Consequently, spatial token-based models tend to require more computational resources to achieve similar performance to that of DDT tokens, which can more efficiently explore the solution space.

\section{Conclusion}
\label{sec:conclusion}
In this paper, we propose the PhysAR framework to generate videos that are consistent with physical laws by integrating symbolic reasoning and reinforcement learning.      
By introducing the Diffusion Timestep Tokenizer (DDT), we overcome the limitations of spatial tokens in video generation, enabling the generation model to perform physical reasoning in a symbolic manner.   
Through the reinforcement learning and reward functions based on physical quantities (\eg, velocity and mass), our model gradually optimizes the physical properties of the generated video, overcoming the limitations of data-driven methods in physical extrapolation.  Experimental results show that the PhysAR framework can generate videos that comply with the physical laws, particularly when facing out-of-distribution situations, where they still effectively discover and adhere to physical laws within dynamic video sequences.

\section*{Acknowledgements}
This work was supported in part by the National Natural
Science Foundation of China (No.62037001, No.62307032), and the ``Pioneer'' and ``Leading Goose'' R\&D Program of Zhejiang under Grant No. 2025C02022.
{
    \small
    \bibliographystyle{ieeenat_fullname}
    \bibliography{main}
}
\clearpage
\maketitlesupplementary

\appendix

\addtocontents{toc}{\protect\setcounter{tocdepth}{2}}
\paragraph{Overview}
In this supplementary material, we present the following content:
\vspace{-3em}
\renewcommand{\contentsname}{}
\tableofcontents
\appendix

\section{Implementation Details}
\subsection{Tokenizer}
\paragraph{Encoder.} 
In the encoder, we set $T=16$ (\ie, the number of query tokens). The dimensions of the query tokens and the noise-free image are $R^{16\times 256}$ and $R^{1024\times 256}$, respectively.
The encoder contains two independent transformers, each comprising $12$ layers with the latent dimension of $128$. 
Following SD3~\cite{esser2024scaling}, despite the noise-free images and query tokens being input into separate transformers, we join the sequences of the two for the attention operation. This allows both representations to operate independently while considering the influence of the other. 
The encoder output retains only the transformed query tokens, serving as the image's latent representations.

\paragraph{Quantizer.} The quantizer is an EMA-variant of vector quantization. Following \cite{yu2021vector}, we leverage a linear projection from the encoder output to low-dimensional variable space for code index lookup (\textit{i.e.}, reduced from a $256$-d vector to a $16$-d vector per code). We also apply L2 normalization to the encoded latent features and codebook latent variables. Moreover, at each training step, we reset the dead entries in the codebook $\mathcal{C}$ (\ie, rarely matched with any tokens) to random tokens in the training batch. 

\paragraph{Decoder.}  We use the same MMDiT architecture proposed in SD3~\cite{esser2024scaling}  for our decoder with minor modifications.
Each transformer in the MMDiT comprises $12$ layers with latent dimension of $512$. 
The sequence of quantized tokens replaces the text tokens as input, with a linear layer to project the $16$-dimensional quantized vector to the latent dimension (\textit{i.e.}, $512$) of the MMDiT. Additionally, we also removed the pooled token embedding introduced in SD3.

Furthermore, the hyper-parameters of training the tokenizer are detailed in Table~\ref{tab:hyper}.

\begin{table*}[t]
    \centering

    \vspace{-1em}
    \begin{tabular}{lccc}
    \toprule
    \textbf{Hyper-parameters} & \textbf{LLM-pre-train} & \textbf{LLM-post-train} & \textbf{Tokenizer}\\ \hline
    LLM init & LLama3.1-8B & LLM-pre-train & - \\
    Optimizer & AdamW & AdamW & AdamW \\
    Optimizer param. & \multicolumn{2}{c}{$\beta_1=0.9,\beta_2=0.95,\epsilon=1\mathrm{e}{-6}$} & $\beta_1=0.9,\beta_2=0.99,\epsilon=1\mathrm{e}{-6}$ \\
    Peak LR & 1e-4 & 3e-6 & 1e-4 \\
    LR scheduler & Cosine & Cosine & Linear+Cosine \\
    Batch size & 768 & 28 & 1024 \\
    Training Steps & 360K & 1600 & 140K \\
    Warmup Steps & 5K & 100 & 5K \\
    Weight decay   & 0.05 & 0.05 & 0.0 \\
    Gradient clipping    & 1.0 & 1.0 & - \\
    Numerical precision    & bfloat16 & bfloat16 & bfloat16 \\
    Resource Usage    & 80 NVIDIA A800 & 8 NVIDIA A800 & 32 NVIDIA A800 \\
    Framework    & DDP & DDP & DDP \\
    \bottomrule
    \end{tabular}
    \caption{\label{tab:hyper}The detailed training hyper-parameters. ``LLM-pre-train'' denotes the pretraining of 
    Phys-AR, ``LLM-post-train'' denotes the reinforcement tuning of Phys-AR, while ``Tokenizer'' denotes the training of DDT.}
    \vspace{-1em}

\end{table*}

\section{Tokenizer Reconstruction}
\begin{table*}[]
\centering
\begin{tabular}{ccccc}
\toprule
\multicolumn{1}{c}{Scenario} & \multicolumn{1}{c}{Ground Truth Error} & \multicolumn{1}{c}{VAE Error (DiT)} & \multicolumn{1}{c}{VQGAN Error} & \multicolumn{1}{c}{DDT Error} \\
\hline
Uniform               & 0.0099                                 & 0.0105                                             & 0.0118                                         & 0.0121                                        \\
Collision                    & 0.0117                                 & 0.0131                                             & 0.0155                                         & 0.0147                                        \\
Parabola                     & 0.0210                                 & 0.0210                                             & 0.0231                                         & 0.0198                                    \\
\bottomrule
\end{tabular}
\caption{Comparison of reconstruction errors among different methods, including ground truth videos, VAE-reconstructed videos, VQGAN-reconstructed videos, and Diffusion Timestep Tokenizer (DDT)-reconstructed videos. This comparison aims to evaluate the accuracy of each approach in preserving the video content, assessing their ability to encode and decode visual information.}
\label{tab:rec}
\end{table*}
In this paper, we propose a Diffusion Timestep Tokenizer, which aims to recover visual attributes lost during the diffusion process by learning discrete, recursive visual tokens, thereby enhancing the accuracy of video reconstruction. The core idea of this approach is to leverage timestep information from the diffusion process to compensate for the loss of critical visual features in conventional visual encoding methods. To ensure the fairness and comparability of experimental results, we evaluate and compare the reconstruction errors of three different tokenizers across three types of scene videos in our dataset. These tokenizers include a VAE pre-trained encoder on real images (used for visual encoding in DiT), a spatial token-based VQGAN that we trained on Phyworld videos, and the proposed Diffusion Timestep Tokenizer, which also trained on Phyworld videos.

In our experiments, Ground Truth Error is used to measure the discrepancy between real videos and the parameters of the physical simulator while also accounting for the errors of the VAE encoder in the DiT model. To further assess the performance of different tokenizers, we compute the video reconstruction errors for VQGAN and the Diffusion Timestep Tokenizer, respectively. As shown in Table~\ref{tab:rec}, the results indicate that the reconstruction errors of all three tokenizers are highly similar to the Ground Truth Error, demonstrating their capability to accurately encode and decode the physical event videos used in this study.

\section{More Dataset Details}
\subsection{Fundamental Physical Scenarios Data}
For the Box2D simulator, PhyWorld~\cite{kang2024far} initializes the world as a 10 $\times$ 10 grid, with a timestep of 0.1 seconds, resulting in a total time span of 3.2 seconds (32 frames). For all scenarios, they set the radius $r \in [0.7, 1.5]$ and velocity $v \in [1, 4]$ as in-distribution (in-dist) ranges. Out-of-distribution (OOD) ranges are defined as $r \in [0.3, 0.6] \cup [1.5, 2.0]$ and $v \in [0, 0.8] \cup [4.5, 6.0]$.

\subsection{Collision Scenario}
The four degrees of freedom (DoFs) are the masses of the two balls and their initial velocities, fully determining the collision outcomes. PhyWorld contains 3k, 30k, and 3M training samples by sampling grid points from the 4-dimensional in-dist joint space of radii and velocities. For in-dist evaluation, they randomly sample about 2k points from the grid, ensuring the data are not part of the training set. For OOD evaluation, they sample from the OOD ranges, generating approximately 4.8k samples across six OOD levels: (1) only $r_1$ OOD, (2) only $v_1$ OOD, (3) both $r_1$ and $r_2$ OOD, (4) both $v_1$ and $v_2$ OOD, (5) $r_1$ and $v_1$ OOD, and (6) $r_1$, $v_1$, $r_2$, and $v_2$ OOD. Additionally, for collisions, they ensure that all collisions occur after the 4th frame in each video, allowing the initial velocities of both balls to be inferred from the conditioned frames.

\subsection{Uniform and Parabolic Motion}

The two DoFs are the ball’s mass and initial velocity. They contain 3k, 30k, and 3M training samples by sampling from the 2-dimensional in-dist joint space of radius and velocity. For in-dist evaluation, they sample approximately 1.05k for uniform motion and 1.1k for parabolic. For OOD evaluation, they generate about 2.4k (uniform motion) and 2.5k (parabolic) samples across three OOD levels: (1) only $r$ OOD, (2) only $v$ OOD, and (3) both $r$ and $v$ OOD.

For all scenarios, we follow PhyWorld to filter out videos where the ball exits the field of view prematurely by inspecting the simulation state, as these videos do not provide sufficient meaningful information for the model.

\section{Detail of the Reward Function}
We introduce a reward function to convert the velocity error to the excitation signal:
\begin{equation}
R_{\text{vel}}=\alpha e^{-k\text{v}_{error}}
\end{equation}
where $\alpha$ controls the maximum reward and $k$ determines the reward decay rate, allowing for a differentiated response to various error intervals. 

\begin{figure}[t]
\centering
\includegraphics[width=0.45\textwidth]{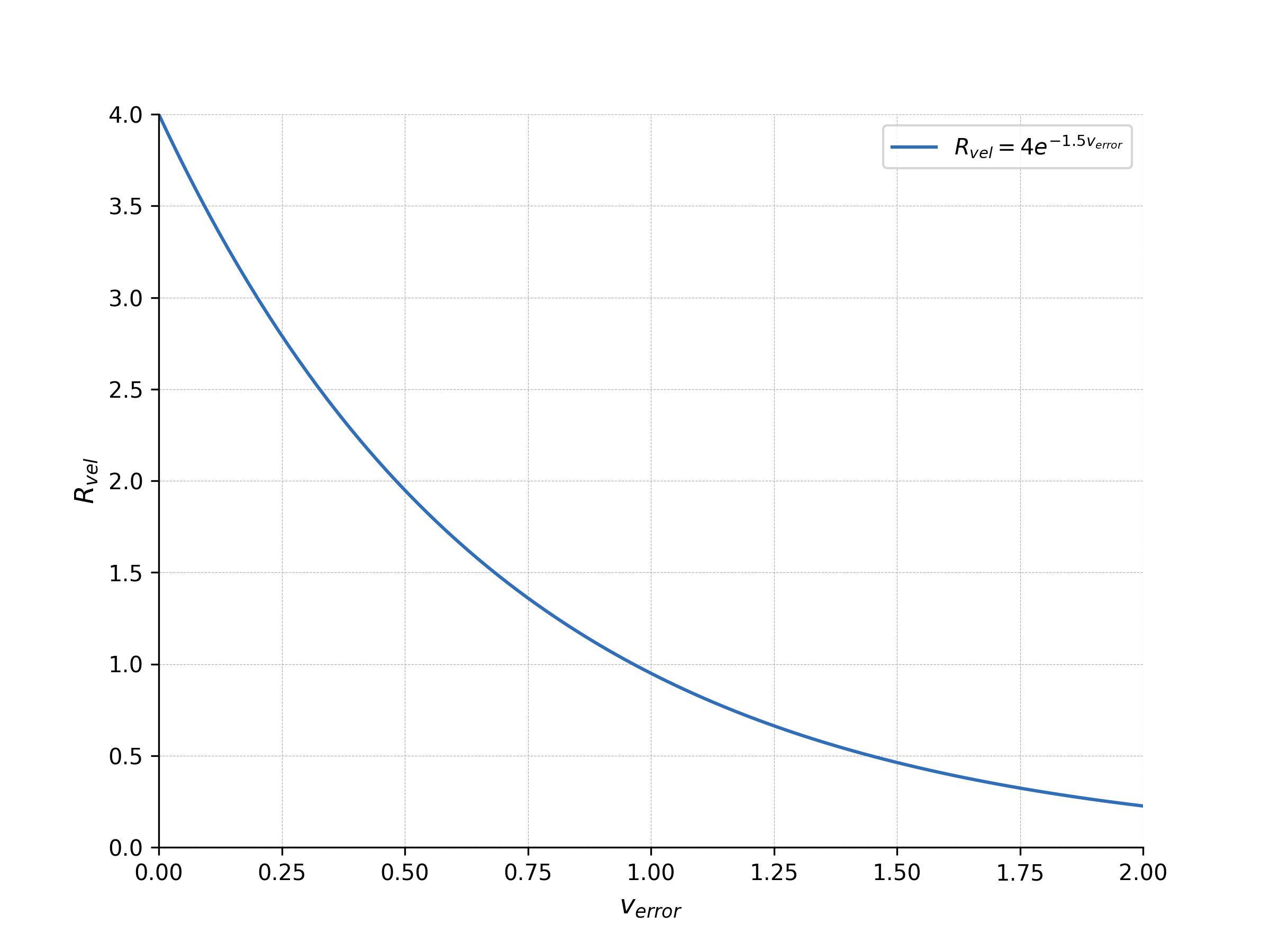}
\vspace{-1em}
\caption{Visualization of the velocity reward function as the velocity error changes.}
\vspace{-1.5em}
\label{fig:r_vel}
\end{figure}

In practice, we set $\alpha=4$ and $k=1.4$. The variation of the reward signal with error is shown in Figure~\ref{fig:r_vel}.
This nonlinear mapping strategy creates a reward space with a steep gradient in the low error region ($\text{v}_{error} \rightarrow 0$), enhancing the model’s sensitivity to accurate predictions.

\section{More Visualization Examples}
\begin{figure*}[t]
\centering
\includegraphics[width=\textwidth]{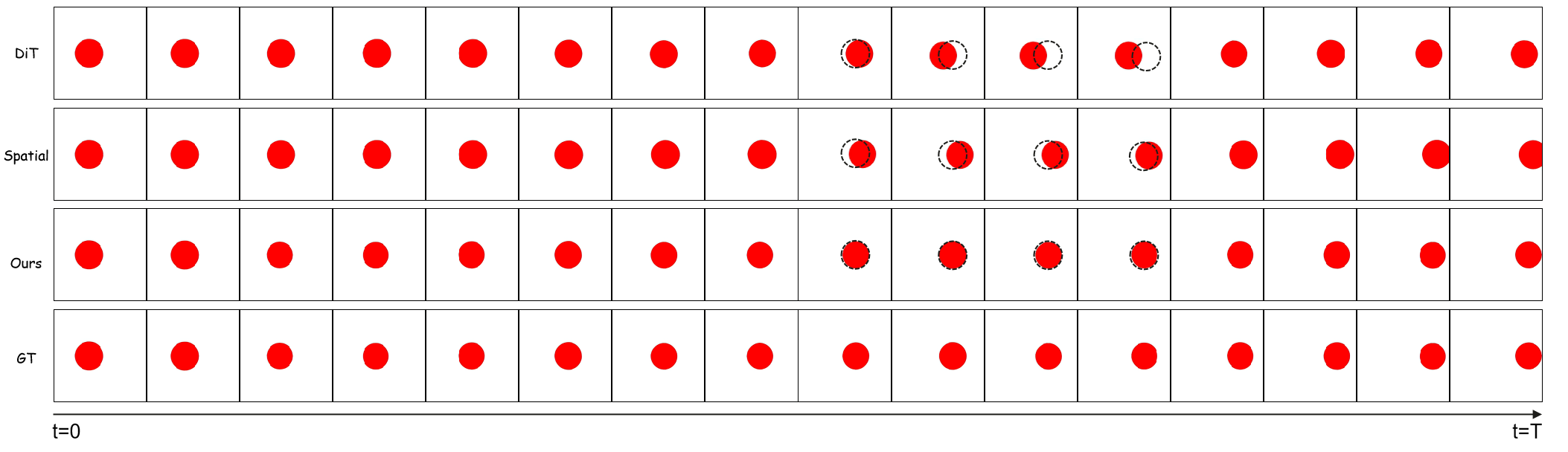}
\vspace{-1em}
\caption{Visualization of uniform motion.}
\vspace{-1.5em}
\label{uni}
\end{figure*}

\begin{figure*}[t]
\centering
\includegraphics[width=\textwidth]{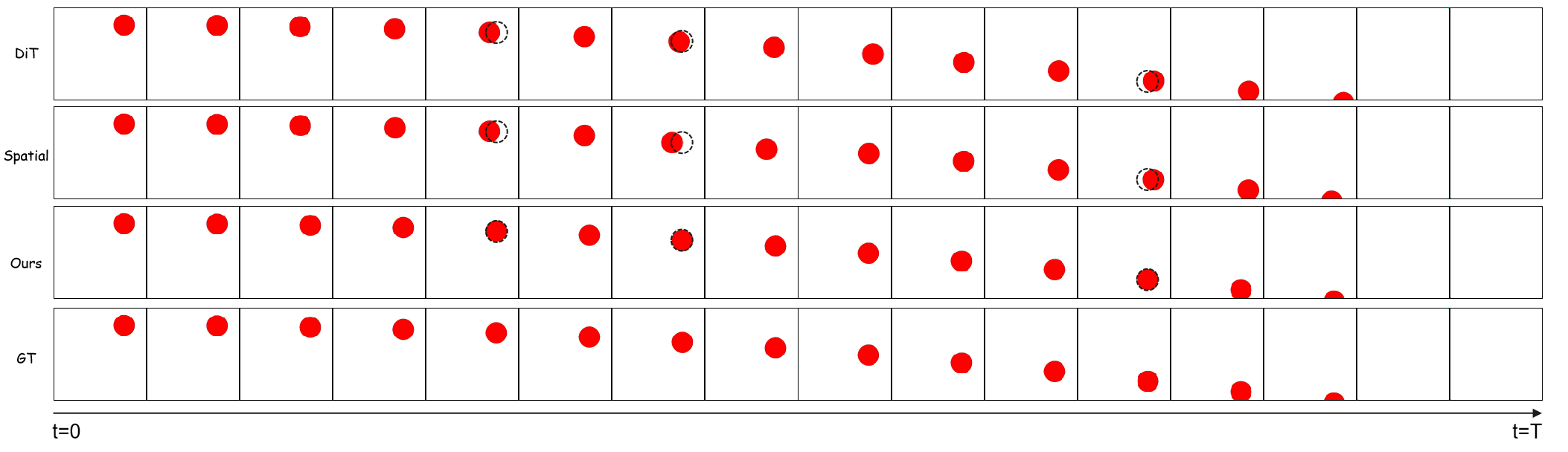}
\vspace{-1em}
\caption{Visualization of parabola motion.}
\vspace{-1.5em}
\label{par}
\end{figure*}

\begin{figure*}[t]
\centering
\includegraphics[width=\textwidth]{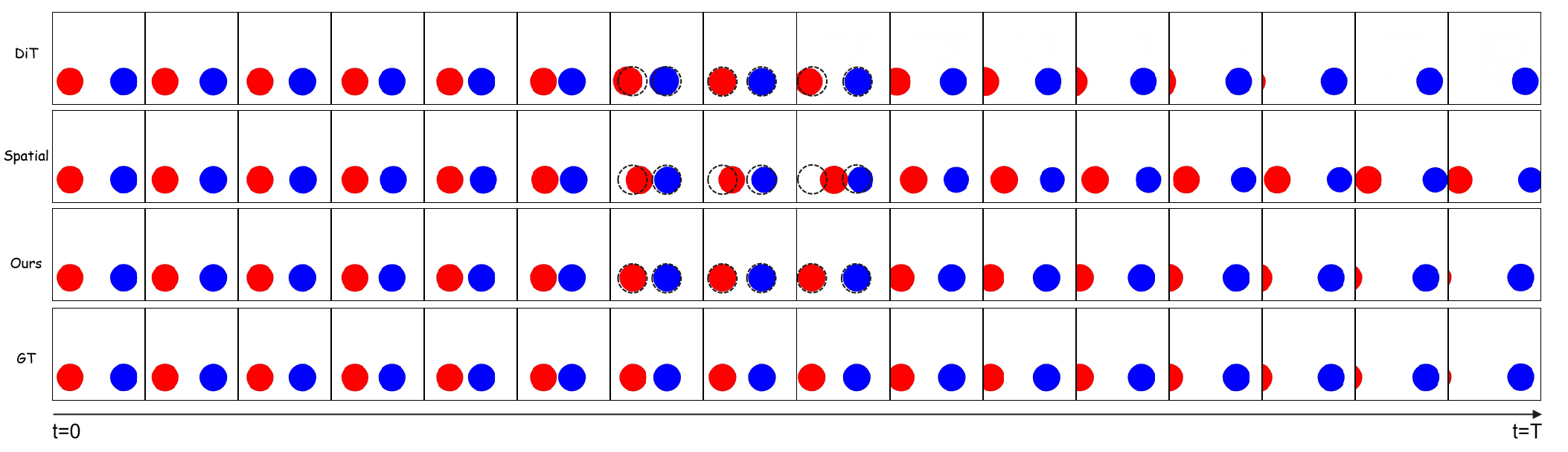}
\vspace{-1em}
\caption{Visualization of collision motion.}
\vspace{-1.5em}
\label{col}
\end{figure*}

We have incorporated visualizations of three types of uniform motion (Figure~\ref{uni}), parabolic (Figure~\ref{par}), and collision (Figure~\ref{col}) different models to further evaluate the effectiveness of our approach. These visualizations provide an intuitive understanding of how each tokenizer captures and reconstructs the underlying motion dynamics. By analyzing the reconstructed trajectories, we can assess the ability of different models to preserve spatial and temporal consistency, accurately depict motion patterns, and recover fine-grained details lost during encoding.

\newpage

\newpage

\section{Limitation and Future Work }
\label{sec:future work}
Based on the diffusion timestep tokenizer, we employed an autoregressive generative model to successfully predict videos of three types of simulated physical motions (uniform motion, parabolic motion, and collisions) during the post-training phase of reinforcement learning. Although real-world videos involve more complex physical parameters and dynamics, which pose challenges on the design of reward functions and generative models, we believe this work presents a promising technical approach for the large-scale generation of physical videos and scalable world models. We intend to explore this direction further as a focus of future research.

We are currently working on improving and scaling up the training of our DDT-tokenizer and the MLLM on a significantly larger dataset (about 500M images). 
\textit{\textbf{In the near future, we will release a more powerful version, along with a detailed technical report. Stay tuned!}}
Building on this foundation, we aim to demonstrate further that visual reinforcement learning is a significant approach for addressing more visual-language tasks such as image comprehension~\cite{wu2024semantic,wang2024instruction,huang2024autogeo,pan2024auto,pan2024towards,yan2024low,yan2025diff,jin2024rethinking,guo2025efficient}, video comprehension~\cite{lin2023tavt, lin2023exploring, emotion,li2023multi,wang2023weakly,wang2023semantic,guo2025bridging} and text-to-image generation~\cite{lin2024non, linaction,pan2025generativemultimodalpretrainingdiscrete,wang2025towards}.

\end{document}